\documentclass[nohyperref]{article}

\usepackage{microtype}
\usepackage{graphicx}
\usepackage{booktabs} %
\usepackage{multirow}
\usepackage{caption}
\usepackage{subcaption}
\usepackage{titlesec}

\usepackage[accepted]{icml2022}

\usepackage{amsmath}
\usepackage{amssymb}
\usepackage{mathtools}
\usepackage{amsthm}
\usepackage{xspace}

\usepackage{hyperref}
\usepackage[capitalize,noabbrev]{cleveref}
\usepackage[usestackEOL]{stackengine}
\usepackage{tikz}
\theoremstyle{plain}

\theoremstyle{definition}

\theoremstyle{remark}

\usepackage[textsize=tiny]{todonotes}

\renewcommand{\paragraph}[1]{
     \textbf{#1} 
 }

\begin{document}

\twocolumn[
\icmltitle{Coder Reviewer Reranking for Code Generation}

\icmlsetsymbol{equal}{*}

\begin{icmlauthorlist}
\icmlauthor{Tianyi Zhang${}^{*}$}{stanford}
\icmlauthor{\enspace Tao Yu}{hku}
\icmlauthor{\enspace Tatsunori B. Hashimoto}{stanford}
\icmlauthor{\enspace Mike Lewis}{fair}
\icmlauthor{\enspace Wen-tau Yih}{fair}
\icmlauthor{\enspace Daniel Fried}{cmu}
\icmlauthor{\enspace Sida I. Wang}{fair}
\end{icmlauthorlist}

\icmlaffiliation{fair}{Meta AI - FAIR}
\icmlaffiliation{stanford}{Stanford University}
\icmlaffiliation{cmu}{Carnegie Mellon University}
\icmlaffiliation{hku}{The University of Hong Kong}

\icmlcorrespondingauthor{Tianyi Zhang}{tz58@stanford.edu}
\icmlcorrespondingauthor{Sida I. Wang}{sida@meta.com}

\icmlkeywords{Machine Learning, ICML}

\vskip 0.3in
]

\printAffiliationsAndNotice{\icmlEqualContribution} %

\newcommand{\ensuretext}[1]{#1}
\newcommand{\marker}[2]{\ensuremath{^{\textsc{#1}}_{\textsc{#2}}}}
\newcommand{\arkcomment}[3]{\ensuretext{\textcolor{#3}{[#1 #2]}}}
\newcommand{\tianyi}[1]{\arkcomment{\marker{T}{Z}}{#1}{blue}}
\newcommand{\dfried}[1]{\arkcomment{\marker{D}{F}}{#1}{olive}}
\newcommand{\mbr}[1]{MBR-\textsc{Exec}}
\newcommand{\ie}[0]{\emph{i.e.}\xspace}
\newcommand{\eg}[0]{\emph{e.g.}\xspace}

\begin{abstract}
Sampling diverse programs from a code language model and reranking with model likelihood is a popular method for code generation but it is prone to preferring degenerate solutions.
Inspired by collaborative programming, we propose Coder-Reviewer reranking.
We augment \emph{Coder} language models from past work, which generate programs given language instructions, with \emph{Reviewer} models, which evaluate the likelihood of the instruction given the generated programs.
We perform an extensive study across six datasets with eight models from three model families.
Experimental results show that Coder-Reviewer reranking leads to consistent and significant improvement (up to 17$\%$ absolute accuracy gain) over reranking with the Coder model only.
When combined with executability filtering, Coder-Reviewer reranking can often outperform the minimum Bayes risk method.
Coder-Reviewer reranking is easy to implement by prompting, can generalize to different programming languages, and works well with off-the-shelf hyperparameters.
\end{abstract}

\section{Introduction}
Recent pretrained language models (PLMs) have demonstrated an impressive ability to generate code given natural language instructions~\citep{codex, incoder, palm, codegen}.
One popular technique is to use a generative language model trained on code, which we call the Coder model, to sample multiple code solutions for a single instruction and rerank the solutions based on the likelihood the Coder model assigns to each~\cite{codex}.
Despite its wide-spread use, reranking with the Coder model often mistakenly prefers degenerate solutions, \emph{e.g.}, extremely short code or repetitive solutions.
As a result, reranking performance often decreases when the number of candidate program increases~(\Cref{fig:num_samples_teaser}).
These biases are known to arise in language models when using mode-seeking inference methods such as greedy decoding~\citep{curious-case} or beam search~\citep{mmi-diversity,stahlberg2019nmt}.

In this work, we take inspiration from collaborative software development.
For example, in the standard practice of code review, programmers submit implementations given specifications and have the submitted code cross validated by other code reviewers.
We instantiate this idea by using prompting to obtain a Reviewer model, which checks the generated programs against the language instruction. 
Formally, we first sample programs $y$ given the instruction $x$ via the Coder model $p(y|x)$ and cross check via the Reviewer model $p(x|y)$.
The Reviewer model reinforces the language instruction by evaluating the likelihood of every word in the instruction.
\begin{figure}
    \centering
    \includegraphics[width=1.00\linewidth]{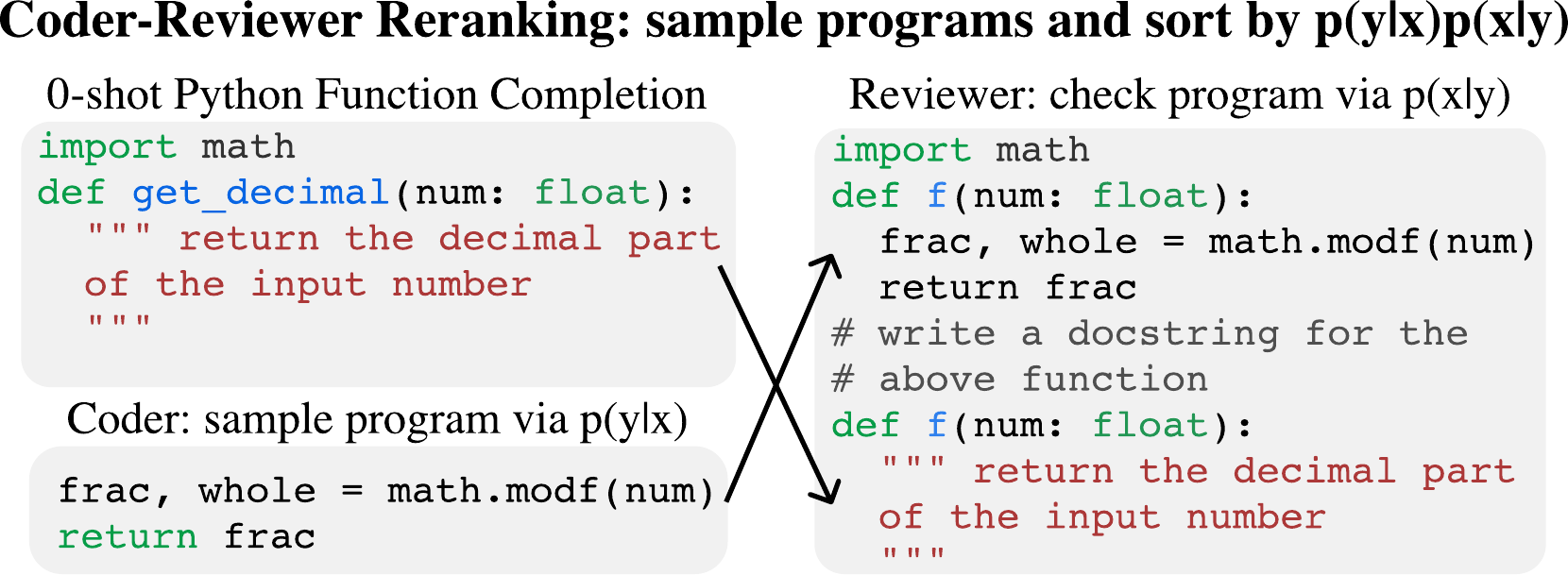}
    \caption{Given a language instruction $x$, a Coder model samples programs $y$, and a Reviewer model checks the generated programs against the instruction by measuring $p(x|y)$. Coder-Reviewer reranking solicits a consensus between Coder and Reviewer by ranking with their product $p(x|y)p(y|x)$.}
    \label{fig:teaser}
    \vspace{-15pt}
\end{figure}

To obtain a consensus between the Coder and the Reviewer, we propose Coder-Reviewer reranking which selects the solutions by the product of the reviewer model and the coder model, $p(x|y)p(y|x)$.
We show that Coder-Reviewer is a specific instantiation of the Maximum Mutual Information (MMI) objective~\citep{mmi-diversity}, which favors solutions that have high mutual information with the instruction and down weights generic solutions (where $p(y)$ is high).
MMI has also been shown to be effective against degenerate solutions in many other natural language processing tasks~\citep{rerank-parsing,generative-qa, pragmatic-inference}.

\looseness=-1 To implement the Reviewer model $p(x|y)$, we propose a simple prompting method. %
After a program $y$ is generated by the Coder model $p(y|x)$, we invert the order in which the instruction $x$ and the solution $y$ appear in the prompt, and query the pretrained language model again to estimate $p(x|y)$.
Our prompting approach avoids any additional training and is easy to generalize to different programming languages.
\Cref{fig:teaser} shows an example prompt: the Coder model generates programs given the function header and the docstring; we then extract the generated program and place it before the docstring when prompting the Reviewer model.

We carry out an extensive empirical study on six datasets with three different programming languages and experiment with seven models from three model families. 
Compared to past works' approach of ranking with the Coder model $p(y|x)$ alone, Coder-Reviewer reranking demonstrates consistent and effective performance gains (up to $17\%$ absolute accuracy gain).
When combined with executability filtering, Coder-Reviewer reranking can often outperform the minimum Bayes risk decoding method~\citep{mbr}, which involves more complex aggregation of the executed outputs.
The code is available on GitHub.\footnote{\url{https://github.com/facebookresearch/coder_reviewer_reranking}}
\section{Related Work}
\paragraph{Code Generation.}
Many prior works explored code generation with neural networks~\citep{suggest-method-name, ling-etal-2016-latent, concode, syntactic-code, program-repair} and many benchmarks have been proposed to evaluate code model performance~\citep{apps, spider, nl2bash}.
Recently, large language models pretrained on code have shown unprecedented zero-/few-shot ability, and can even perform well in code competitions that are challenging to programmers~\citep{palm, codex,mbpp,alphacode}.
Our work builds on the impressive ability of pretrained code models and achieves additional gains by leveraging a Reviewer model that evaluates the probability of the language instruction given generated programs.

\paragraph{Maximum Mutual Information}(MMI) and its variants have been shown to be effective in many natural language processing tasks, including text classification~\citep{noisy-classification}, speech processing~\citep{Bahl1986MaximumMI}, dialogue~\citep{mmi-diversity}, instruction following~\citep{pragmatic-inference}, question answering~\citep{generative-qa}, and semantic parsing~\citep{rerank-parsing}.
In contrast to Maximum Likelihood which optimizes $\log p(x|y)$, 
MMI optimizes the pointwise mutual information $\log \frac{p(x,y)}{p(x)p(y)}$.
In practice, it is popular to optimize a weighted version of the MMI objective~\cite{mmi-diversity}. We show in \Cref{sec:coder-reviewer} that Coder-Reviewer reranking is a specific instantiation of the weighted MMI objective.
However, Coder-Reviewer reranking differs from this work by
leveraging prompting to obtain the Reviewer model $p(x|y)$, rather than training a separate model, and by showing that the objective produces substantial benefits for the task of code generation. %
Concurrently, \citet{flipped} explore a MMI-like prompting approach for reasoning tasks.

\paragraph{Reranking Methods for Code Generation.}
\citet{codex} point out that the diverse samples from large language models often contain correct programs and they propose to rank program samples by the Coder model $p(y|x)$.
Since then, many methods have been proposed to leverage sample consistency~\citep{mbr} or training supervised rerankers~\citep{fault-aware}
In particular, \citet{mbr} and \citet{alphacode} propose to cluster program surface forms using the executed outputs of the generated programs.
\citet{codet} propose to generate unit tests for Python function completion problems and design selection programs that validate generated programs and unit tests against each other.
On the one hand, Coder-Reviewer reranking does not require execution, which enables it to be applied in more diverse scenarios, and Coder-Reviewer reranking is not specific to any programming languages or packages.
On the other hand, Coder-Reviewer reranking is orthogonal and complementary to methods that incorporate execution semantics:
in \Cref{sec:main-results}, we show empirical results on the benefits of combining Coder-Reviewer reranking with execution-based ranking methods.
\section{Background}

\begin{figure*}[t]
     \centering
     \begin{subfigure}[b]{0.32\textwidth}
         \centering
         \includegraphics[width=\linewidth]{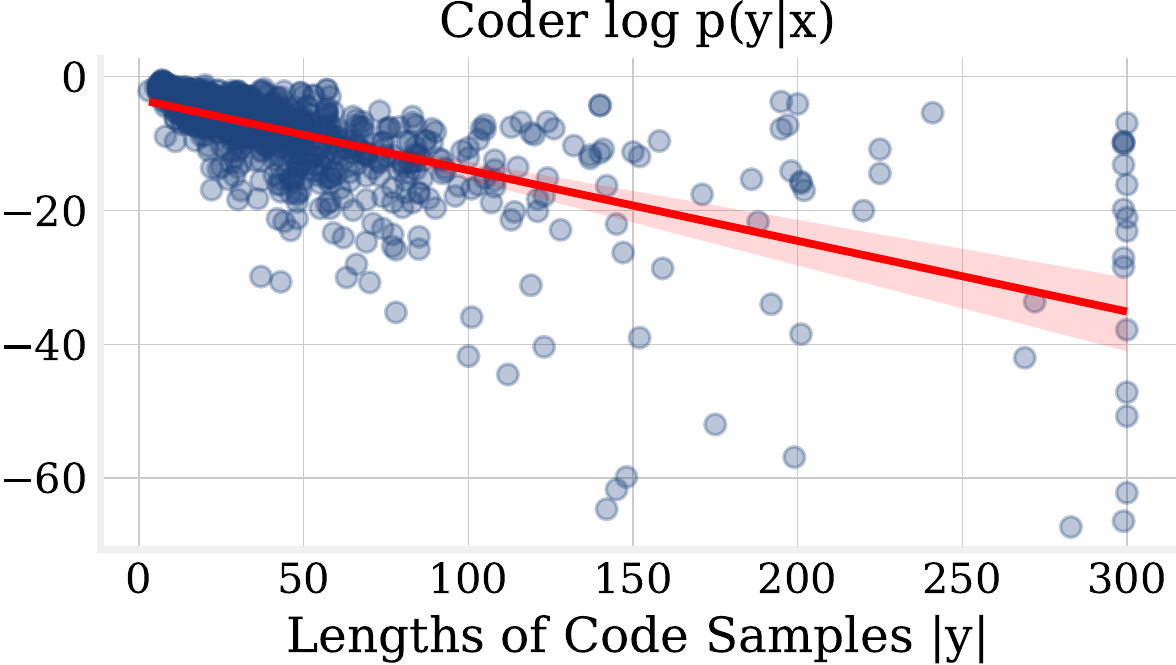}
         \caption{}
         \label{fig:sum_len}
     \end{subfigure}
     \hfill
     \begin{subfigure}[b]{0.32\textwidth}
         \centering
         \includegraphics[width=\linewidth]{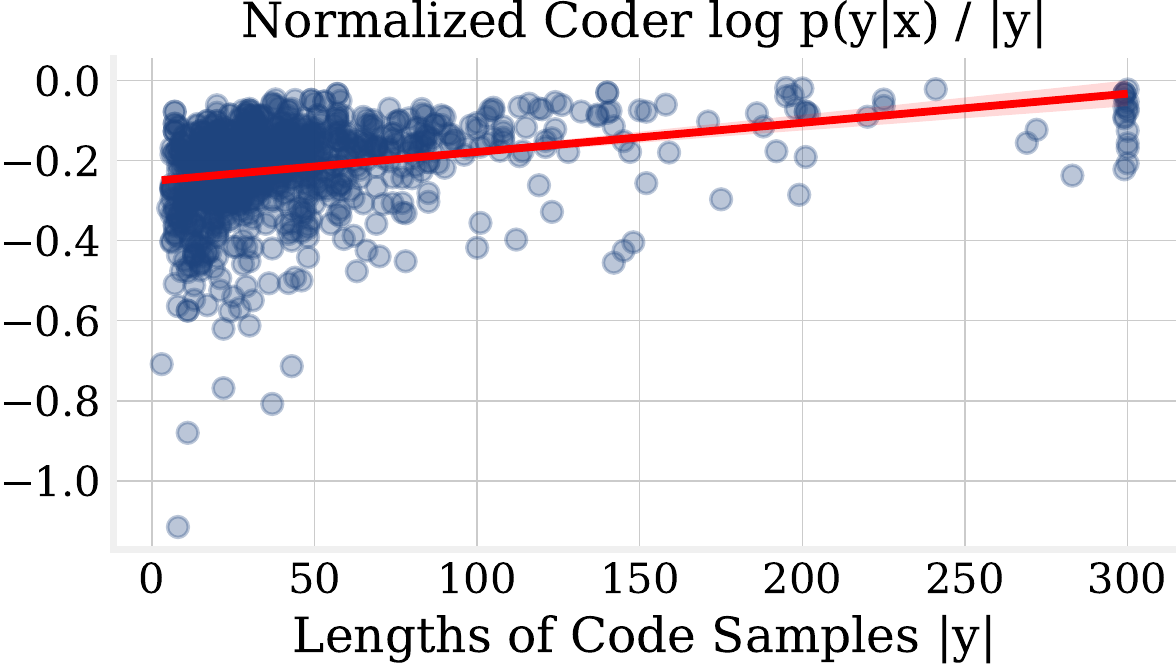}
         \caption{}
         \label{fig:mean_len}
     \end{subfigure}
     \hfill
     \begin{subfigure}[b]{0.32\textwidth}
         \centering
         \includegraphics[width=\linewidth]{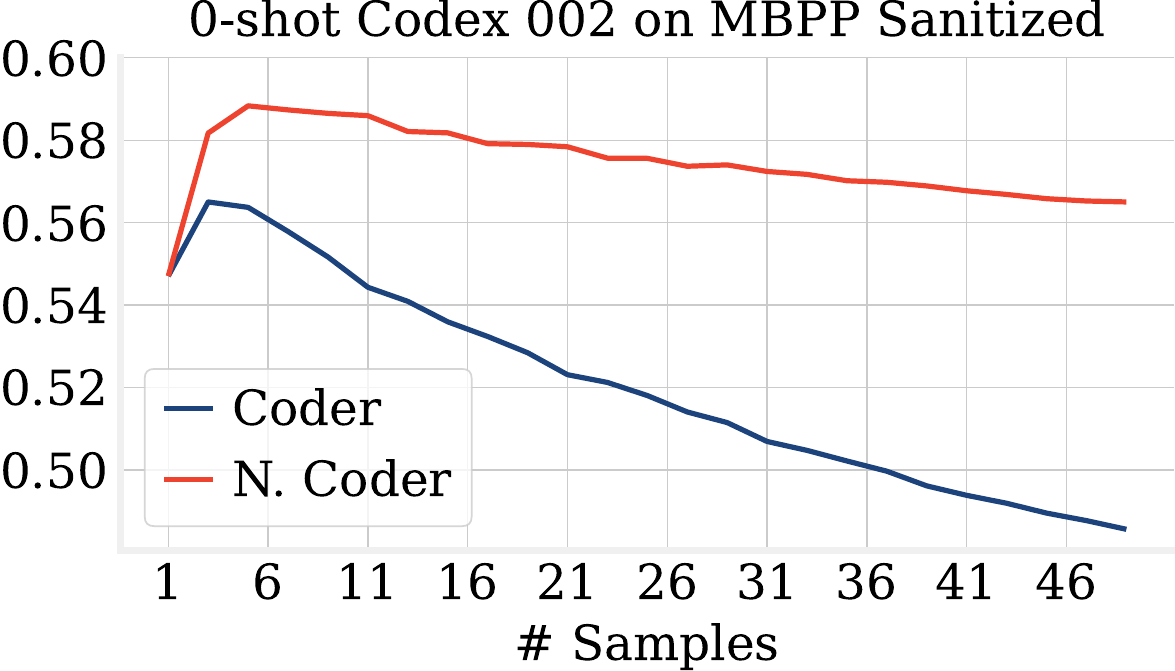}
         \caption{}
         \label{fig:num_samples_teaser}
     \end{subfigure}
     \vspace{-10pt}
     \caption{\ref{fig:sum_len} and \ref{fig:mean_len} show that the Coder model $p(y|x)$ has a strong dependence on the length of generated code and length normalization can introduce additional biases. \ref{fig:num_samples_teaser} shows that in practice, Coder-only reranking (Coder) and normalized Coder-only reranking (N. Coder) have worse performance when the number of samples becomes large, which can be caused by selecting degenerate solutions.}
     \vspace{-15pt}
\end{figure*}

\paragraph{Zero-/Few-shot Code Generation.}
We are interested in using pretrained code language models to generate code $y$ conditioned on natural language instructions $x$.
In addition, we assume access to a context $c$ that provides useful code context such as package imports or data dependencies.
In the example shown in \Cref{fig:teaser}, $x$ is the instruction ``return decimal part of the input number'', $y$ is the generated function body, and $c$ includes importing the math package.
For few-shot code generation, we also have $n$ demonstration examples $(\hat{x}_{1}, \hat{y}_{1}, \hat{c}_{1})$ to $(\hat{x}_{n}, \hat{y}_{n}, \hat{c}_{n})$.

To solve this problem, our key leverage is a Coder model --- a conditional language model $p_{\theta}(y|c,x)$, which can generate programs given the instruction and context.
In this case of few-shot generation, the Coder model also conditions on the demonstration examples, i.e., we consider $p_{\theta}(y | \hat{c}_{1}, \hat{x}_{1}, \hat{y}_{1}, \ldots, \hat{c}_{n}, \hat{x}_{n}, \hat{y}_{n}, c, x )$.
Throughout this work, we implement these conditional models using prompting: placing the relevant objects such as $c$ and $x$ into the input context of the language model.

\paragraph{Collecting Program Samples.}
To obtain programs from the Coder model $p_{\theta}$, we sample autoregressively using a temperature-scaled $p_{\theta}$.
 If the temperature is low, temperature scaling increases the probability of sampling well-formed programs.

\paragraph{Coder-Only Reranking.}
Once we have collected 25-100 samples, we rerank and select a single program as our output program.
A popular technique in prior work \citep{codex} is to pick the program $y$ that has the highest Coder model likelihood $\log p(y|x, c)$.
However, the Coder model likelihood $\log p(y|x, c)$ is biased toward shorter programs~\citep{stahlberg2019nmt}.
To counter this shortcoming, \citet{codex} propose to apply length normalization and instead rerank using the average token-level log likelihood, $\frac{1}{\lvert y \rvert}\log p(y|c, x)$.

\paragraph{Failures of Coder-Only Reranking.}
\Cref{fig:sum_len} and \Cref{fig:mean_len} plot the Coder $\log p(y|x)$ and Normalized Coder $\frac{1}{\lvert y \rvert}\log p(y|c, x)$ values against the lengths of generated programs $\lvert y \rvert$.
From these figures we can observe that Coder-only reranking prefers short solutions; applying length normalization overcorrects and prefers long solutions.
This is a known issue of reranking/searching using the likelihood of neural language models~\citep{curious-case,stahlberg2019nmt}.
In practice, many of these short programs are trivial (\eg containing only a \texttt{return} or \texttt{pass} statement) and many of these long programs contain repetitive code. %
\Cref{fig:num_samples_teaser} plots the accuracy of ranking with Coder model or the length normalized Coder model (N. Coder) versus the number of program samples.
We observe that accuracy degrades as the number of program samples increases, since degenerate but high-scoring solutions have an increasing chance of appearing as more programs are sampled.
This demonstrates that both types of Coder likelihood are insufficient as reranking objectives for code generation.
We further analyze this behavior in \cref{sec:analysis}.

\section{Coder-Reviewer Reranking}
\label{sec:coder-reviewer}
\paragraph{Reviewer Model $p(y|x)$.}
In real-world professional software development, it is common to have programmers review each other's work.    
Motivated by this, and the previously-demonstrated shortcomings of Coder-only reranking, we introduce a Reviewer model.
Note that we use the same underlying model $p_{\theta}$ for both the Coder and the Reviewer model but prompt differently.
In the case of few-shot code generation, we will have $p_{\phi}(x|c, y, \hat{c}_{1}, \hat{y}_{1}, \hat{x}_{1}, \ldots, \hat{c}_{n}, \hat{y}_{n}, \hat{x}_{n})$.
Compared to the Coder model, the Reviewer model is tasked with evaluating the likelihood of the instruction $x$.
Degenerate programs cannot account for the instruction well and therefore will often have a low Reviewer model likelihood.

\begin{figure}[t]
     \centering
     \includegraphics[width=0.65\linewidth]{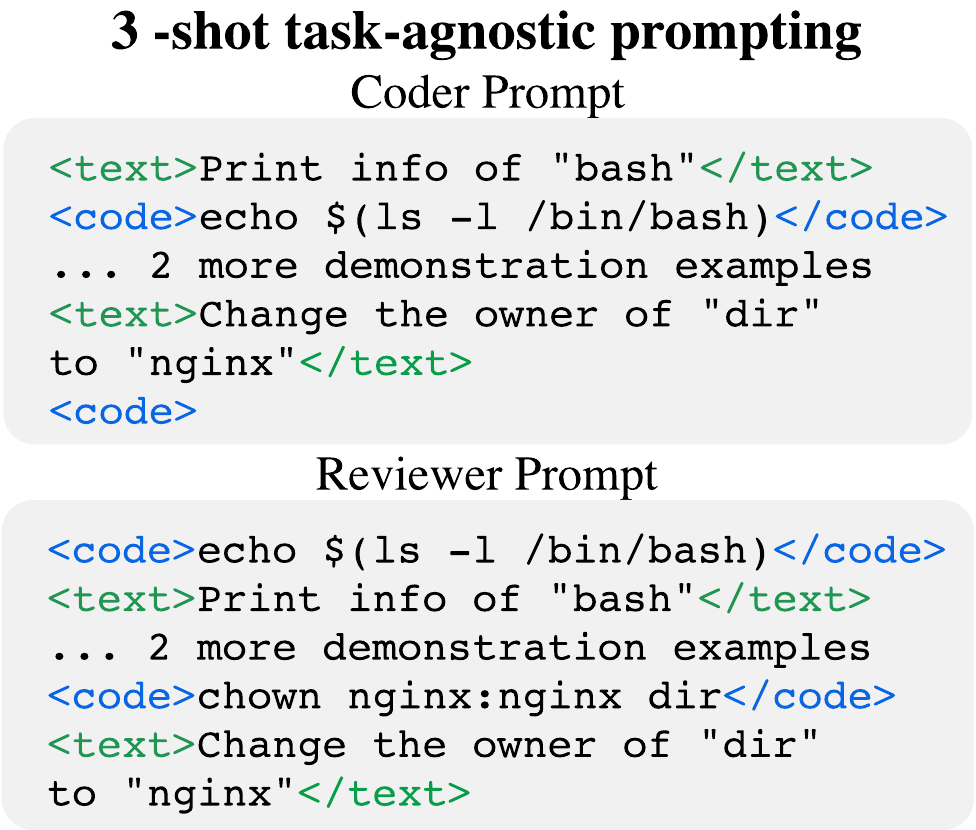}
     \caption{Example task-agnostic prompt on the NL2Bash dataset. We invert the order in which language instruction and the generated program appear to estimate the Reviewer model $p(x|y)$.}
     \label{fig:prompt_example}
     \vspace{-15pt}
\end{figure}

\paragraph{Prompting for the Reviewer Model.}
We adopt a prompting-based approach to implement the Reviewer model.
In short, once the program samples $y$ are generated, we invert the order in which instruction $x$ and program $y$ appear in the input context and query the language model $p_{\theta}$ again to obtain their likelihood.
Recall that in \Cref{fig:teaser}, we demonstrate a task-specific prompt we designed for Python function completion datasets.
In this prompt, we first duplicate the function header and place the generated program $y$ before the instruction docstring $x$.
Additionally, we insert a natural language task specification ``write the docstring for the above function'' into the prompt to further specify the task to the pretrained language model.
In \Cref{fig:prompt_example}, we give another example of a 3-shot task-agnostic prompt on the NL2Bash dataset.
\citet{mbr} propose this task-agnostic Coder model prompt that marks the location of instruction and generated programs with html-like tags.
For prompting the Reviewer model, we swap the language instruction and the generated program, along with their tags.
Importantly, we apply this inversion to both the demonstration examples and the last test example.

\paragraph{Combining Coder and Reviewer.}
To leverage the information provided by both the Coder and the Reviewer model, we propose Coder-Reviewer reranking, which reranks programs using the product of the Coder and the Reviewer, 
\begin{multline*}
\log p(x|y)p(y|x) =
\log p(x|y) + \log p(y|x) \\ \text{(Coder-Reviewer Reranking}).
\end{multline*}
Similarly, we can apply Coder-Reviewer reranking when using the length normalized Coder score,
\begin{multline*}
\frac{\log p(x|y)}{\lvert x \rvert} + \frac{\log p(y|x)}{\lvert y \rvert} \\ \text{(Normalized Coder-Reviewer Reranking}).
\end{multline*}
Here, we normalize the reviewer model score $p(x|y)$ to match the scale of the normalized Coder model score.

Because Coder-Reviewer Reranking combines two models by taking their product, Coder-Reviewer Reranking is sensitive to low probability under either model.
As a result, Coder-Reviewer reranking seeks out program samples that obtain a consensus from both the Coder and the Reviewer.
In the next section, we illustrate that by seeking a consensus, Coder-Reviewer can successfully alleviate the biases from each individual component.

\paragraph{Understanding the Relation between Coder-Reviewer Reranking and Maximum Mutual Information.}
We now show that Coder-Reviewer reranking is a special instantiation of the Maximum Mutual Information objective~\citep{mmi-diversity}. 
To avoid notation clutter, we abstract away the problem structure specific to code generation and denote task input $x$ and output $y$.
The usual Maximum Likelihood objective optimizes $p(y|x)$, which is the same as reranking with the Coder model.
However, Maximum Likelihood has the risk of preferring generic or degenerate answers~\citep{curious-case} and \citet{mmi-diversity} propose to optimize Mutual Information $\frac{p(x, y)}{p(x)p(y)}$.
Moreover, \citet{mmi-diversity} propose to optimize for a weighted version of the Mutual Information objective, using a weighting parameter $\alpha$:
\begin{align}
    & \text{argmax}_{y}\; \log \frac{p(y, x)}{p(x)p(y)^{\alpha}}\\
    =\; &\text{argmax}_{y}\;(1-\alpha)\log p(y|x) + \alpha \log p(x|y) \\
    =\; &\text{argmax}_{y}\; (1-\alpha)\log p(y|x) + \alpha \log \frac{p(y, x)}{p(x)p(y)}
\end{align}
From (2), we 
see that Coder-Reviewer reranking is a special case of this objective where $\alpha$ is set to $0.5$ (we include a derivation in \Cref{sec:app-derivation}).
From (3), we see that both can be viewed as interpolating likelihood with a mutual information criterion, with strength $\alpha$.
In other words, Coder-Reviwer reranking favors program samples that have high mutual information with the language instruction and therefore can filter out low quality solutions that cannot explain the instruction well.
This perspective also motivates exploring using hyperparameter $\alpha$ to control the mixing ratio between the Coder and the Reviewer.
In section~\Cref{sec:app-alpha-tuning}, we show that the off-the-shelf hyperparameter $\alpha=0.5$ usually works well already, and tuning $\alpha$ can give a small additional gain.
Next, we conduct an quantitative analysis to show that Coder-Reviewer model can filter low quality programs.

\section{Analysis of Degenerate Cases}
\label{sec:analysis}

\begin{figure}[]
     \centering
     \includegraphics[width=0.8\linewidth]{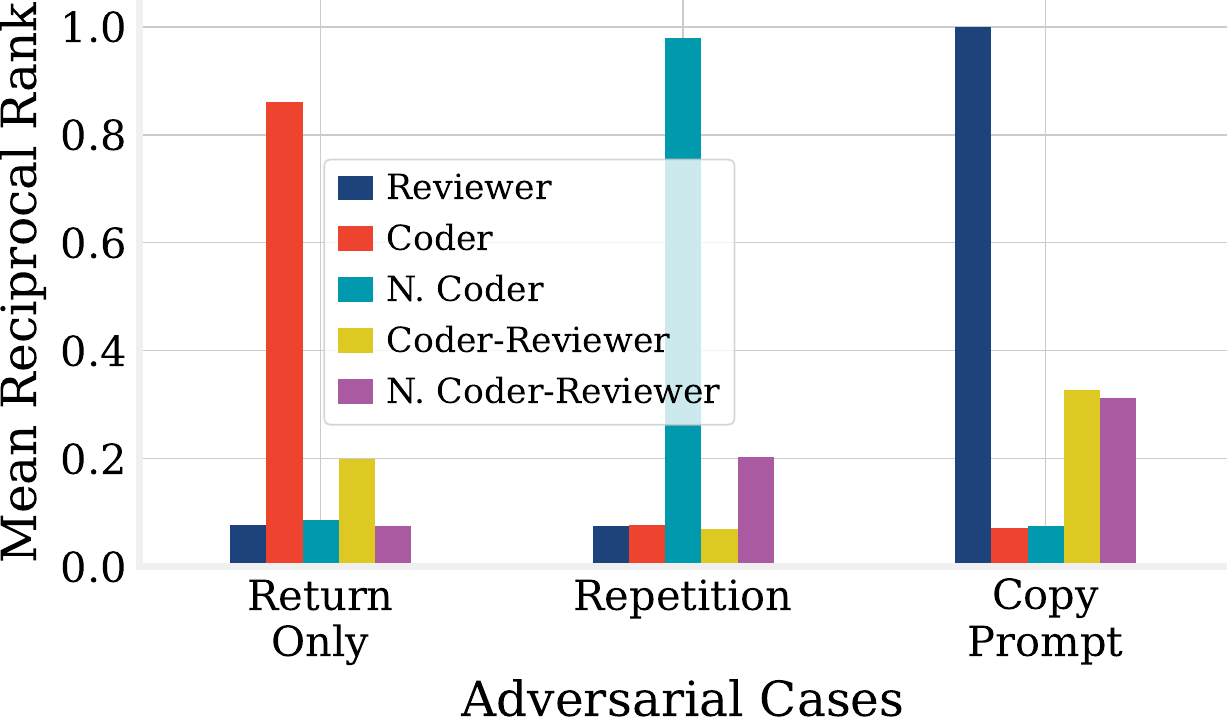}
     \caption{Mean Reciprocal Rank of ranking three typical degenerate cases. Coder Reviewer variants alleviate the biases in its individual components.}
     \label{fig:adv_cases}
     \vspace{-10pt}
\end{figure}
In this section, we experiment with ranking artificially created degenerate programs to analyze the preference of different ranking methods.
We experiment with the Codex002 model~\citep{codex} on the 0-shot sanitized MBPP dataset~\citep[MBPP-S][]{mbpp}, which is a Python function completion dataset with example prompt structure shown in \Cref{fig:teaser}.
We defer experimental details and dataset details to \Cref{sec:exp-setup}.

For each MBPP problem, we sampled 25 programs via the Coder model $p(y|x)$.
Then, we construct three cases that reflect common degeneracies and inject them among the 125 samples.
We construct these cases:
A. \texttt{ReturnOnly} where the function body only contains a return statement 
B. \texttt{Repetitive} where the function body contains print statements printing from 1 to 50, with each line containing exactly one print statement.
C. \texttt{CopyPrompt} where the function body only contains a comment that repeats the language instruction.

With these programs, we now compare the reranking behavior of \underline{Coder} $p(y|x)$, normalized Coder (\underline{N.Coder}) $\frac{1}{\lvert y \rvert}\log p(y|x)$, and \underline{Reviewer} $p(x|y)$.
We also included \underline{Coder-Reviewer} reranking and the normalized version \underline{N.Coder-Reviewer} into the comparison.
\Cref{fig:adv_cases} plots the mean reciprocal rank (MRR) of the three degenerate cases ranked by different methods, where high MRR suggests a bias toward the degenerate case.
\Cref{fig:adv_cases} shows that both Coder and Reviewer methods fail on different degenerate programs: Coder favors programs that are effectively empty, even though those clearly do not follow the language instructions; Normalized Coder favors programs full of repetitions; and Reviewer can be biased toward spurious surface form overlap between generated programs and the language instruction.
In contrast, Coder-Reviewer reranking can alleviate the biases in its individual components.

\paragraph{Degenerate Solutions Rejection.}
Based on this analysis, we prescribe three easy-to-implement procedures to reject degenerate solutions before reranking takes place.
By strengthening the Coder and Reviewer baselines, we know that any performance gain brought by Coder-Reviewer cannot be easily attributed to simple fixes.
\begin{itemize}
    \item We filter empty programs. For Python function completion datasets, we also filter trivial solutions that only contain \texttt{return} or \texttt{pass}.
    \item We filter repetitive programs whose \texttt{zlib} compressed representation is more than four times shorter than the original program.\footnote{This threshold corresponds to \texttt{Repetitive}, which prints from $1$ to $50$.}
    \item For all Python code generation problems, we use an off-the-shelf canonicalization software \texttt{pyminifier} to remove comments, docstrings, and replace all print and assertion messages to empty strings. This procedure helps reduce the spurious surface form overlap with the language instruction. For function completion problems, we also standardize the function names. 
\end{itemize}

We analyze the effects of the above procedure in \Cref{sec:rejection-ablation}.
Overall, we find that these procedures help improve both the baseline Coder methods and our proposed Coder-Reviewer methods. 
Next, we present our extensive empirical study to demonstrate that Coder-Reviewer is an effective and consistent method.

\section{Experimental Setup}
\begin{figure*}[ht]
     \centering
     \includegraphics[width=0.8\linewidth]{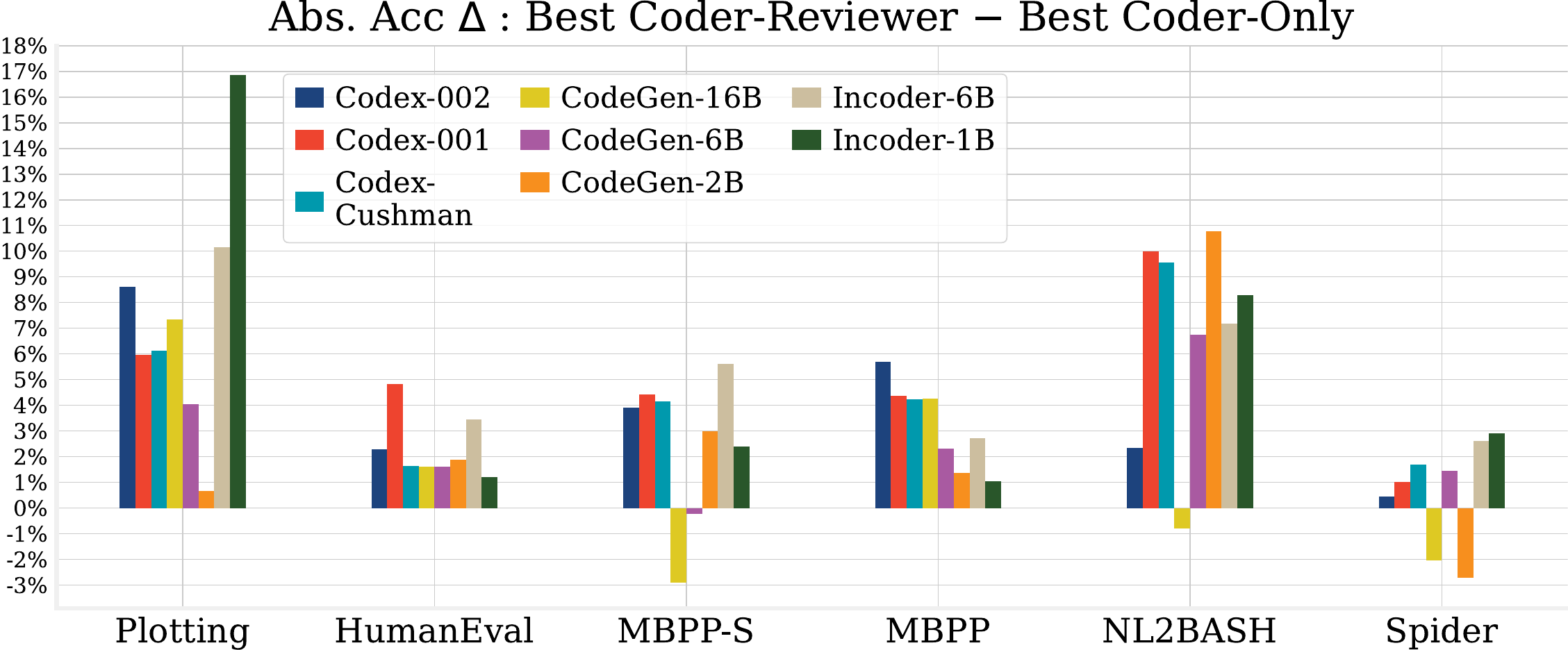}
     \caption{Absolute accuracy difference between the best Coder-Reviewer variants (with or without length normalization) and the best Coder variants (with or without length normalization). We observe performance gain from applying Coder-Reviewer on 43 out of 48 dataset cross model pairs.}
     \label{fig:acc_delta}
     \vspace{-13pt}
\end{figure*}

\label{sec:exp-setup}
We present an extensive empirical study that spans six datasets and three model families.
Our evaluation data consists of both zero-shot and few-shot settings and investigates both task-specific and task-agnostic prompt design, and we study eight models whose parameter counts range across two orders of magnitude.

\subsection{Datasets}
We provide a brief description of our experimental datasets and present a detailed description in \Cref{sec:app-dataset-details}.
We first consider three zero-shot datasets, which all involve generating Python code.
\textbf{HumanEval} and \textbf{MBPP-Sanitized} are two popular Python function completion datasets.
On these datasets, we generally follow the prompt design in \citet{codet} and use the Reviewer prompt presented in \Cref{fig:teaser}.
\paragraph{Plotting} is a subset of DS-1000~\citep{Lai2022DS1000} and contains $155$ realistic questions adapted from StackOverflow about \texttt{matplotlib}. 
The rest of DS-1000 cannot be easily handled by left-to-right autoregressive because it often introduces additional contexts that are better addressed by infilling models.
We leave applying Coder-Reviewer reranking to infilling models for future work.

Next, we consider three three-shot datasets, \textbf{MBPP}, \textbf{Spider}, and \textbf{NL2Bash}, which include programming languages other than Python.
Our task-agnostic prompts and other experimental design on these three datasets are all taken from \citet{mbr} and are similar to the one presented in \Cref{fig:prompt_example}.

\subsection{Models}
\paragraph{Codex~\citep{codex}} models are descendants of GPT-3~\citep{gpt3}.
We consider three variants of the codex model: Codex002 (\texttt{codex-davinci-002}), Codex001 (\texttt{codex-davinci-001}), and Codex-Cushman (\texttt{codex-cushman}).
While the exact modeling details are not made public, we assume based on their codenames that Codex002 and Codex001 are $175$B models and Codex-Cushman is a $12$B model.

\paragraph{InCoder~\citep{incoder}} models are trained with the causal mask language modeling objective and are capable of both infilling and left-to-right generation.
In this work, we only leverage their autoregressive ability to keep the comparison consistent with other model families and leave the investigation of its infilling ability for future work.
We experiment with $6$B and $1$B models from this family.

\paragraph{CodeGen~\citep{codegen}} is a family of autoregressive language models pretrained on code data. We experiment with the $16$B, $6$B, and $2$B CodeGen models.

\subsection{Implementation Details}
Due to the lack of validation split on the benchmarks we experimented with, we restrain ourselves from hyperparameter search and rely on a single set of decoding hyperparameters.
On all datasets, we sample with temperature $0.4$ and set the max tokens to be $300$.
For our main results in \Cref{tab:main_table}, we sample $125$ different programs for each problem and then bootstrap $50$ times to report the mean accuracy of reranking $25$ samples.
Given the small size of these datasets, we find bootstrapping to be helpful in reducing variance.
In \Cref{sec:hyperparameters}, we analyze the impact of decoding hyperparameters on the validation accuracy.
We apply the proposed degenerate solution rejection to all baseline and proposed methods and study its effect in \Cref{sec:rejection-ablation}.
Additionally, we apply executability filtering to all baseline and proposed methods to better compare with state-of-the-art methods that rely on execution more heavily.
Executability filtering~\citep{mbr} removes programs that produce runtime errors before applying any other ranking methods.
We present the ranking results without executability filtering in \Cref{sec:app-all-results}.

\subsection{Reranking Methods}
\paragraph{Baselines.} We first compare to baseline selection methods that are popular in existing work. \underline{Random} reports the percentage of correct programs after degenerate solution rejection.
\underline{Coder} selects the program that has the highest $p(y|x)$ and \underline{N. Coder} applies length normalization to the Coder likelihood.

\paragraph{Proposed Methods.}
We then compare the methods proposed in this work. \underline{Reviewer} selects programs that has the highest $p(x|y)$.
\underline{Coder-Reviewer} ranks via the product of the Coder and the Reviewer model scores and \underline{N. Coder-Reviewer} applies length normalization to the component model scores before taking the product.

\paragraph{State-of-the-art Methods.}
We compare to the competitive Minimum Bayes Risk method~\citep[\mbr{};][]{mbr}, which compares the executed outputs of the generated programs on test inputs and selects programs whose outputs are most typical.
While for Python function completion problems it is straightforward to compare execution outputs, this is less obvious when execution results involve more complex objects.
On the plotting dataset, due to the complexity and the multimodal nature of \texttt{matplotlib}\footnote{plots with different styles such as colors, fonts, etc. might all be correct but it is difficult to aggregate them.}, we compare the output figures directly pixel-by-pixel.
On HumanEval, we extract the first test case appearing in the docstring and use the executed output for \mbr{}.
For other datasets, we follow the practices recommended in \citet{mbr}.
We compare to \mbr{} using the same setting with baseline and proposed methods in \Cref{tab:main_table}.

We also compare to CodeT~\citep{codet}, which generates assertions for Python function completion problems and proposes a novel graph-based aggregation process for selecting programs based on program-test agreement.
While CodeT is a competitive method, it relies on the ability to generate good assertions and cannot be easily used on other programming languages/tasks.
For example, models typically cannot generate unit tests used in the Plotting dataset because good tests involve complicated assertions on the figure object and do not naturally appear in public data.
Another example is the spider dataset, where generating additional unit tests involves creating different input database~\citep{active-prior}, a challenging problem on its own.
Therefore, we separate CodeT from the rest of the comparison and compare to its reported numbers in \Cref{tab:codt_table}.

\section{Experimental Results}
In \Cref{sec:main-results}, we show that Coder-Reviewer reranking is either the best performing method or competitive with the best performing method across all settings.
In \Cref{sec:hyperparameters} and \Cref{sec:rejection-ablation}, we perform ablation studies for the different decisions made in developing Coder-Reviewer reranking and show that Coder Reviewer can work well with default hyperparameters and is more stable than Coder reranking.

\begin{table*}[]
\setlength{\tabcolsep}{1.5pt}
\centering
\resizebox{\textwidth}{!}{%
\begin{tabular}{lccccccccc}
\toprule
{} & \multicolumn{3}{c}{Plotting} & \multicolumn{3}{c}{HumanEval} & \multicolumn{3}{c}{MBPP-S} \\
\cmidrule(lr){2-4}\cmidrule(lr){5-7}\cmidrule(lr){8-10}
{} &            Codex002 &           InCoder6B &          CodeGen16B &            Codex002 &           InCoder6B &          CodeGen16B &            Codex002 &           InCoder6B &          CodeGen16B \\
\midrule
Random            &              $57.6$ &              $21.9$ &              $36.7$ &              $49.2$ &              $16.0$ &              $32.2$ &              $58.8$ &              $25.4$ &              $45.7$ \\
Reviewer          &  $\underline{65.2}$ &     $\mathbf{35.5}$ &     $\mathbf{46.5}$ &  $\underline{61.2}$ &              $17.6$ &              $37.3$ &              $59.5$ &              $28.8$ &              $43.6$ \\
\midrule
Coder             &              $57.7$ &              $15.5$ &              $38.6$ &              $45.1$ &              $20.1$ &              $33.5$ &              $59.8$ &              $28.1$ &  $\underline{50.3}$ \\
Coder-Reviewer    &              $58.3$ &              $16.4$ &              $41.7$ &              $56.7$ &     $\mathbf{23.5}$ &     $\mathbf{40.0}$ &     $\mathbf{64.4}$ &     $\mathbf{33.7}$ &              $47.4$ \\
\midrule
N.Coder           &              $59.4$ &              $22.6$ &              $37.4$ &              $60.2$ &              $19.2$ &              $38.5$ &              $60.5$ &              $27.1$ &              $48.6$ \\
N.Coder-Reviewer  &     $\mathbf{68.0}$ &  $\underline{32.8}$ &  $\underline{45.9}$ &     $\mathbf{62.5}$ &  $\underline{22.0}$ &  $\underline{39.6}$ &              $61.6$ &  $\underline{31.1}$ &              $45.8$ \\
\midrule
MBR-\textsc{Exec} &              $60.9$ &              $21.0$ &              $37.0$ &              $50.5$ &              $20.7$ &              $35.8$ &  $\underline{63.9}$ &              $30.9$ &     $\mathbf{53.5}$ \\
\midrule
{} & \multicolumn{3}{c}{MBPP} & \multicolumn{3}{c}{NL2BASH} & \multicolumn{3}{c}{Spider} \\
\cmidrule(lr){2-4}\cmidrule(lr){5-7}\cmidrule(lr){8-10}
{} &            Codex002 &           InCoder6B &          CodeGen16B &            Codex002 &           InCoder6B &          CodeGen16B &            Codex002 &           InCoder6B &          CodeGen16B \\
\midrule
Random            &              $58.1$ &              $19.6$ &              $39.3$ &              $60.0$ &              $49.8$ &              $35.7$ &              $65.2$ &              $29.4$ &              $25.6$ \\
Reviewer          &     $\mathbf{66.9}$ &              $24.4$ &              $44.1$ &  $\underline{63.3}$ &  $\underline{55.3}$ &              $28.1$ &              $67.5$ &              $38.4$ &              $28.8$ \\
\midrule
Coder             &              $60.2$ &              $23.4$ &              $41.5$ &              $57.4$ &              $48.8$ &              $31.7$ &              $74.1$ &              $38.9$ &     $\mathbf{33.7}$ \\
Coder-Reviewer    &  $\underline{66.4}$ &  $\underline{26.1}$ &  $\underline{46.2}$ &              $61.9$ &              $55.0$ &  $\underline{37.0}$ &  $\underline{74.5}$ &     $\mathbf{41.5}$ &  $\underline{31.7}$ \\
\midrule
N.Coder           &              $60.7$ &              $20.2$ &              $41.9$ &              $61.3$ &              $41.5$ &     $\mathbf{37.8}$ &              $69.9$ &              $38.2$ &              $31.1$ \\
N.Coder-Reviewer  &              $66.2$ &              $24.1$ &              $45.4$ &     $\mathbf{63.7}$ &     $\mathbf{55.9}$ &              $29.5$ &              $71.0$ &  $\underline{40.3}$ &              $29.9$ \\
\midrule
MBR-\textsc{Exec} &              $63.0$ &     $\mathbf{26.7}$ &     $\mathbf{47.3}$ &              $57.4$ &              $48.8$ &              $32.4$ &     $\mathbf{75.2}$ &              $38.2$ &              $30.6$ \\
\bottomrule
\end{tabular}

}
\vspace{-5pt}
\caption{Bootstrapped reranking results with 25 samples. 
Bolded numbers indicate the best results on each column and Underlined numbers indicate the second best results.
In each subsection, we compare including or not including Reviewer reranking (Random vs. Reviewer, Coder vs. Coder-Reviewer, etc.).
Coder-Reviewer variants mostly outperform Coder variants, and often outperforms the competitive MBR-\textsc{Exec} method.
Across all columns except one, a Coder-Reviewer variant is either the best or the second best method.
}

\label{tab:main_table}
\vspace{-10pt}
\end{table*}

\subsection{Primary Results}
\label{sec:main-results}

First, we observe that Coder-Reviewer variants consistently improves over Coder variants. 
\Cref{fig:acc_delta} plots the absolute accuracy difference between the best Coder-Reviewer variant (with and without length normalization) and the best Coder variant.
Overall, we observe that Coder Reviewer leads to a consistent and significant improvement with very few exceptions.
In certain settings, \eg for InCoder models on the plotting dataset, Coder-Reviewer leads to more than $10\%$ improvement.
The only 5 exceptions out of 48 model cross dataset pair where Coder-Reviewer is worse than Coder all involve CodeGen models.
We suspect that difference in pretraining can cause a difference in the capability to estimate the Reviewer model $p(x|y)$ using the prompt we employ in this work.

Second, in \Cref{tab:main_table}, we present the ranking results on the largest models from each model family along with comparison to other methods.
From \Cref{tab:main_table} we find a Coder-Reviewer variant to be the best method most of the time and even when it is not, it is often the second best method.
Analyzing across all data, we find that Coder-Reviewer variants are the best method $62.5\%$ of the time, Reviewer $12.5\%$ of the time, and MBR-\textsc{Exec} $20.83\%$ of the time.
However, Reviewer and MBR-\textsc{Exec} both have scenarios where they trail behind Coder-Reviewer significantly.
For example when applied to the Codex002 model, MBR-\textsc{Exec} is more than $10\%$ worse than normalized Coder-Reviewer on HumanEval and $8\%$ worse on Plotting.
In practice, we find that MBR-\textsc{Exec} works better when the quality of the test inputs that it executes on is high and when it is easy to aggregate the execution outputs.
Compare HumanEval and MBPP-S, which have very simlar format, MBR-Exec performs much better on MBPP-S than on HumanEval.
We ascribe this failure to the lower quality of the test inputs in HumanEval.

\begin{table*}[]
\setlength{\tabcolsep}{2.2pt}
\centering
\footnotesize
\begin{tabular}{lcccccc}
\toprule
{} & \multicolumn{3}{c}{HumanEval} & \multicolumn{3}{c}{MBPP-S} \\
\cmidrule(lr){2-4}\cmidrule(lr){5-7}
{} &  Codex002 & InCoder6B & CodeGen16B & Codex002 & InCoder6B & CodeGen16B \\
\midrule
CodeT   &    $65.8$ &    $20.6$ &     $36.7$ &   $67.7$ &    $34.4$ &     $49.5$ \\
\midrule
Coder-Reviewer       &    $57.9$ &    $24.3$ &     $42.6$ &   $64.7$ &    $35.8$ &     $50.3$ \\
N. Coder-Reviewer &    $66.9$ &    $22.9$ &     $40.5$ &   $61.0$ &    $30.2$ &     $46.1$ \\
\bottomrule
\end{tabular}
\caption{Bootstrapped reranking results with $100$ samples. CodeT numbers are cited from \citet{codet} and not strictly comparable: Coder-Reviewer variants use the first unit test in the docstring for executability filtering whereas CodeT generated its own unit tests.
That being said, this comparison shows that Coder-Reviewer can achieve strong performance on the Python function completion datasets while being easier to generalize to other language/packages.
}
\vspace{-10pt}
\label{tab:codt_table}
\end{table*}

Third, \Cref{tab:codt_table} shows the performance of CodeT and Coder-Reviewer in the same setting of reranking $100$ samples on HumanEval and MBPP-S.
Here the results of CodeT are not strictly comparable to those of Coder-Reviewer.
Recall that CodeT generates additional test cases for these Python function completion problems so CodeT does not execute any test cases provided in the input docstring.
In contrast, we follow the setup in \citet{mbr} and execute the first test case to perform executbaility filtering.
Still, \Cref{tab:codt_table} shows that Coder-Reviewer can achieve very competitive performance while being a more general method that can be applied to datasets where generating test cases is difficult.

In summary, our experimental results show that Coder-Reviewer reranking consistently improves performance over Coder-only reranking from past work across a diverse set of model families and datasets.
In the following sections, we carry out ablation studies to understand the impact of design decisions and sensitivity to hyperparameters.

\subsection{Understanding Hyperparameters}

\begin{figure}[t]
     \centering
     \includegraphics[width=0.8\linewidth]{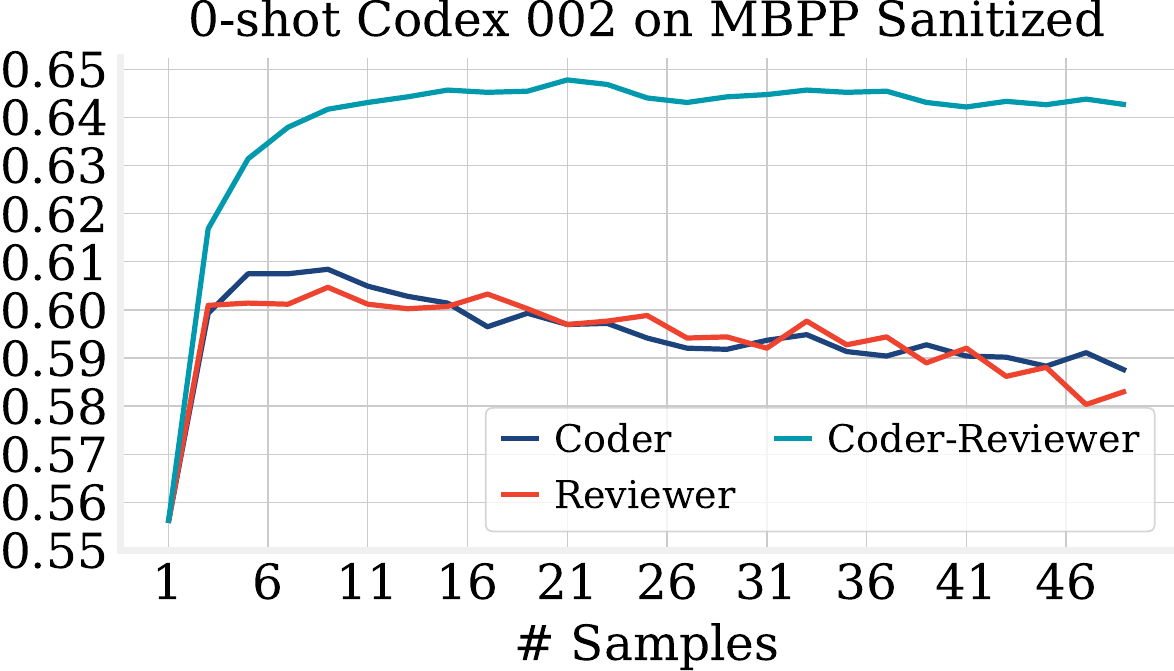}
     \caption{Accuracy versus number of ranking samples. Coder-Reviewer is more stable and robust to degenerate solutions than its individual components.}
     \vspace{-5pt}
     \label{fig:num_examples_final}
     \vspace{-10pt}
\end{figure}

\label{sec:hyperparameters}
We explore the effect of different hyperparameters, including the number of samples and the mixing ratio $\alpha$ between Coder and Reviewer.
We start by plotting the ranking accuracy of Coder, Reviewer, and Coder-Reviewer on the 0-shot MBPP-S dataset with Codex002 in \Cref{fig:num_examples_final} (\Cref{fig:num_examples_appendix} in \Cref{sec:app-samples} includes plots on more methods and datasets).
We observe that in contrast to its components Coder and Reviewer, Coder-Reviewer is more stable in ranking more sample programs, suggesting that it is more robust against potential degenerate solutions when compared to ranking with Coder or Reviewer model only.

In addition, recall that we can introduce a hyperparameter  to control the mixing ratio $\alpha$ between Coder and Reviewer and have the objective $(1-\alpha)\log p(x|y) + \alpha\log p(y|x)$.
In Coder-Reviewer reranking,  we fix $\alpha$ at $0.5$ for simplicity although the weighted objective is also popular in prior work~\citep{mmi-diversity}.
From \Cref{fig:tuning_alpha} in \Cref{sec:app-alpha-tuning}, we observe that the improvement from tuning $\alpha$ is typically less than $1\%$, with a few exceptions with the CodeGen models where tuning can lead to more significant improvements.
Finally, we explore a alternate formulation of the Maximum Mutual Information objective and show that it performs generally worse than Coder-reviewer reranking in \Cref{sec:app-alternate-formulation}.

\subsection{Understanding Degenerate Solution Rejection}
\label{sec:rejection-ablation}
\begin{table}[]
\footnotesize
\centering
\begin{tabular}{lll}
\toprule
{} &               HumanEval &                  MBPP-S \\
\midrule
Random               &           $48.0_{-1.2}$ &           $58.1_{-0.7}$ \\
Coder                &           $38.1_{-7.1}$ &           $55.3_{-4.5}$ \\
N.Coder              &           $59.7_{-0.5}$ &           $60.0_{-0.5}$ \\
Reviewer             &           $57.7_{-3.5}$ &           $55.8_{-3.7}$ \\
Coder-Reviewer       &           $53.2_{-3.5}$ &           $60.5_{-3.9}$ \\
Norm. Coder-Reviewer &  $\mathbf{61.5}_{-1.0}$ &  $\mathbf{60.8}_{-0.7}$ \\
\bottomrule
\end{tabular}

\caption{Ranking results of Codex002 without applying degenerate solution rejection. Numbers in subscripts showing the performance degradation compared to applying programmatic rejection.
We see that degeneration solution rejection is helpful for all methods and unnormalized Coder benefits from the rejection most significantly.
}
\vspace{-15pt}
\label{tab:ablation_table}
\end{table}

We start our analysis by removing the proposed degenerate solution rejection on Codex002's generated programs on HumanEval and MBPP-S.
\Cref{tab:ablation_table} shows the ranking results along with the performance degradation compared to applying the proposed rejection.
We observe that degenerate solution rejection improves all of the baselines and Coder-Reviewer variants, with the most significant effect on Coder.
Importantly, we see that the improvement from rejection is larger on Coder, Reviewer, and Coder-Reviewer than on Random, suggesting that these ranking methods have biases toward the degenerate solutions that we reject.
On these two datasets, the effect of rejection is less obvious on normalized Coder and normalized Coder Reviewer but we find the rejection to be important for these methods on NL2Bash where it is more likely to have egregious repetitions in the program samples.
Finally, these ablation results suggest that Coder-Reviewer still provides a benefit on top of degenerate solution rejection.
Overall, we recommend the usage of our rejection methods because they are easy to implement and can effectively remove the most egregious degeneration programs targeting each ranking method.

\section{Conclusion}
We propose Coder-Reviewer reranking for code generation, which leads to a consistent and significant improvement over the Coder only reranking proposed in prior work.
When combined with executability filtering, Coder-Reviewer reranking can often outperform the \mbr{} method.
Coder-Reviewer is easy to implement with prompting, can generalize to different programming languages, and works well with off-the-shelf hyperparameter.

\bibliography{example_paper}
\bibliographystyle{icml2022}
\appendix
\section{Understanding the Relation between Coder-Reviewer
Reranking and Maximum Mutual Information.}
\label{sec:app-derivation}
We include the derivation from \citet{mmi-diversity} to show that Coder-Reviewer reranking is a special instantiation of the Maximum Mutual Information objective.
We can show
\begin{align*}
    & \text{argmax}_{y}\; \log \frac{p(y, x)}{p(x)p(y)^{\alpha}}\\
    =\; &\text{argmax}_{y}\;(1-\alpha)\log p(y|x) -\alpha\log p(x) + \alpha \log p(x|y)  \\
    =\; &\text{argmax}_{y}\;(1-\alpha)\log p(y|x) + \alpha \log p(x|y)  \\
    =\; &\text{argmax}_{y}\; (1-\alpha)\log p(y|x) + \alpha \log \frac{p(y, x)}{p(x)p(y)},
\end{align*}
by using the fact that $p(x)$ is a constant and removing/adding it does not change the optimziation objective.
\label{sec:app-derivation}

Alternatively, \citet{mmi-diversity} propose another way to instantiate this objective,
\begin{align*}
    & \text{argmax}_{y}\; \log \frac{p(y, x)}{p(x)p(y)^{\alpha}}\\
    =\; &\text{argmax}_{y}\;\log p(y|x) -\alpha\log p(y).  \\
\end{align*}
We show in \Cref{sec:app-alternate-formulation} that this alternate formulation leads to worse performance, which is a similar conclusion to the original finding in \citet{mmi-diversity}.
\section{Additional Experimental details}
\subsection{Detailed Dataset Descriptions}
\label{sec:app-dataset-details}
\paragraph{HumanEval~\citep{codex}} contains $164$ hand-written Python programming questions~\citet{codex}. We use the prompts released by \citet{codet}, which removes the input-output cases present in the original prompts. 
This dataset evaluates the generated function by multiple assertions on the input-output relations.

\paragraph{MBPP-Sanitized~\citep{mbpp}} contains $427$ crowd-sourced Python programming questions. \citet{codet} adapts each problem into having a function header and relocate the natural language instructions into function docstrings, similar to the format of HumanEval. We use this prompt format for our experiment.

\paragraph{MBPP~\citep{mbpp}} original version consists of $974$ python questions, with $500$ of them used for testing and the rest for few-shot prompting.
The prompt taken from \citet{mbr} uses one input-output assertion as additional context to help make the language instruction more specific.
Unlike the 0-shot sanitized version of MBPP, this 3-shot setting requires the models to define the desired function name on its own.

\paragraph{Spider~\citep{spider}} is a benchmark of natural language to SQL query generation.
There are $7000$ examples for training/demonstration and $1034$ questions for testing.
The prompt taken from \citet{mbr} prepends the database schema as the program context.
The generated SQL commands are evaluated with execution accuracy, comparing the database return values to the ones queried by ground-truth SQL commands.

\paragraph{NL2Bash~\citep{nl2bash}} is a benchmark of translating natural language to bash commands.
Because it is difficult to obtain executable enviroments for bash commands, this dataset evaluates character-level BLEU-4 score.

\begin{figure*}[ht!]
     \centering
     \includegraphics[width=\linewidth]{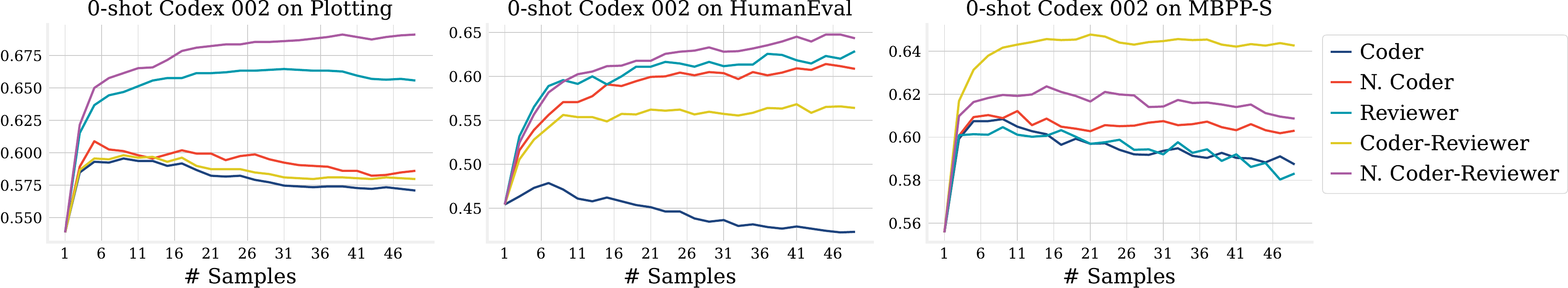}
     \caption{Accuracy versus number of ranking samples. Coder-Reviewer is more stable and robust to degenerate solutions than its individual components.}
     \label{fig:num_examples_appendix}
     \vspace{-5pt}
\end{figure*}

\begin{figure*}[ht!]
     \centering
     \includegraphics[width=0.8\linewidth]{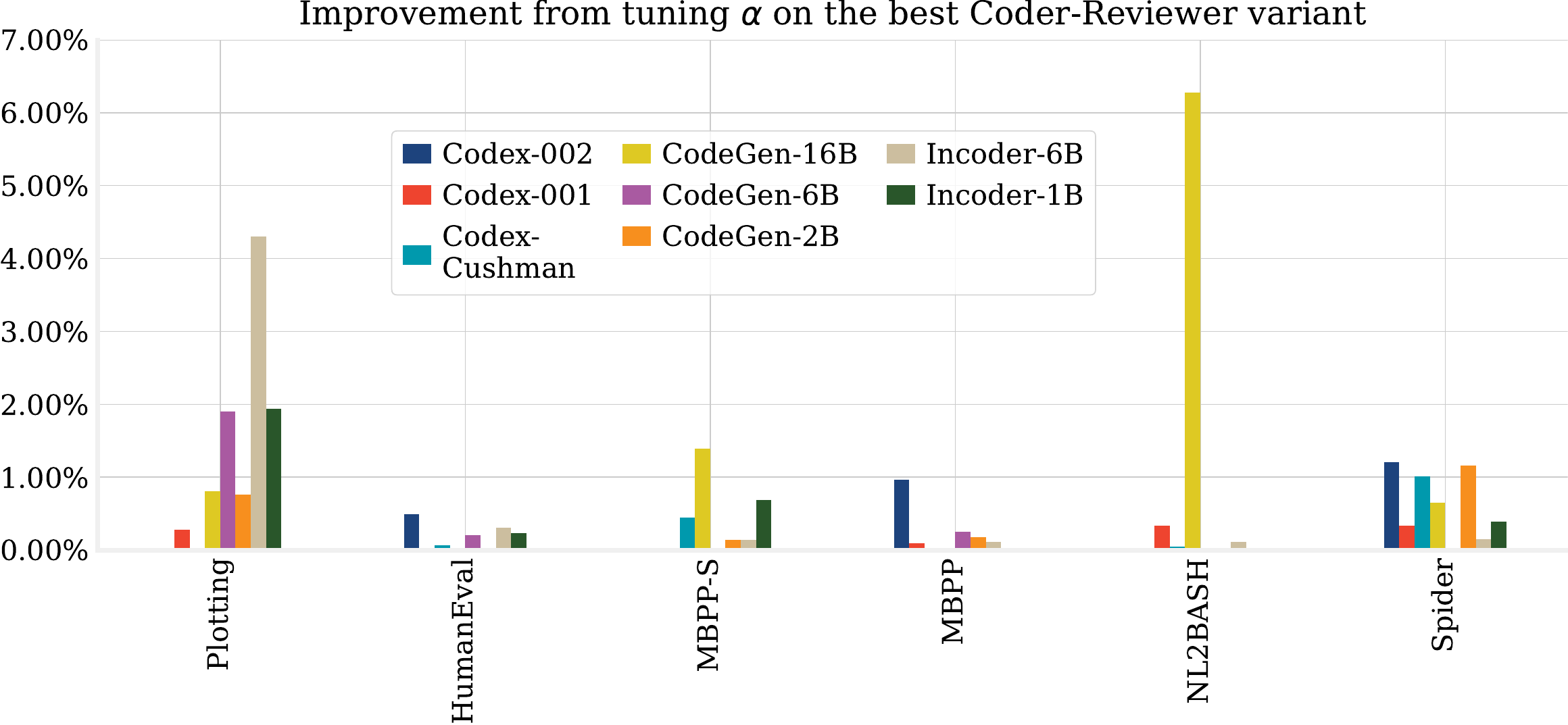}
     \caption{Accuracy improvement from grid searching ensemble mixing ratio $\alpha$. Setting $\alpha=0.5$ (i.e., Coder-Reviewer) usually performs well already and further grid searching mostly only lead to a $1$-$2\%$ improvement. In few settings (CodeGen-$16$B on NL2Bash and Incoder-$6$B on Plotting), tuning $\alpha$ leads to a significant improvement.}
     \label{fig:tuning_alpha}
     \vspace{-5pt}
\end{figure*}

\section{Additional Results}
\label{sec:app-results}
\subsection{Results from All Models}
\label{sec:app-all-results}
\Cref{tab:app_plotting_exec} to \Cref{tab:app_spider_exec} plot ranking results on all eight models we experimented with and it also shows the results of not applying executability filtering.
Analyzing across all data, we find that a Coder-Reviewer variant is the best method $62.5\%$ of the time, Reviewer $12.5\%$ of the time, and MBR-\textsc{Exec} $20.83\%$ of the time.
Notably, a Coder-Reviewer variant is almost always the best method for Codex models with only two exceptions (Codex002 on Spider and MBPP).
Even when none of Coder-Reviewer variants is the best method, a Coder-Reviewer variant is mostly the second best method.
Overall, executability filtering improves most methods and does not change the comparison between different ranking methods.

\subsection{Analyzing the Effect of Number of Program Samples}
\label{sec:app-samples}
\Cref{fig:num_examples_appendix} plots the performance of Coder-Reviewer variants, Coder variants, and Reviewer on all three 0-shot datasets we experimented with.
Similar to \Cref{fig:num_examples_final}, \Cref{fig:num_examples_appendix} shows that compared to Coder variants, Coder-Reviewer variants perform more consistently across the number of program samples being reranked.
Here we plot the performance after applying degeneration solutions rejection.
Still, on all three datasets, the performance of Coder starts to degrade after five or ten samples.
Normalized Coder is more stable on HumanEval and MBPP-S but also does worse with more program samples on Plotting.

\subsection{Analyzing the Effect of Tuning Mixing Ratio $\alpha$}
\label{sec:app-alpha-tuning}

In addition, recall that we can introduce a hyperparameter to control the mixing ratio, $\alpha$ between Coder and Reviewer and have the objective $(1-\alpha)\log p(x|y) + \alpha\log p(y|x)$.
Coder-Reviewer reranking is equivalent to fixing $\alpha$ at $0.5$ although the weighted objective is also popular in prior work~\citep{mmi-diversity}.
To explore the benefits of tuning this additional hyperparameter of $\alpha$, we grid search between $\{0.1, 0.2, \ldots, 0.9\}$ across all experimental settings.
From \Cref{fig:tuning_alpha} in \Cref{sec:app-alpha-tuning}, we observe that the improvement from tuning $\alpha$ is typically less than $1\%$, with a few exceptions (Incoder-$6$B on Plotting and CodeGen-$16$B on NL2Bash) where tuning can lead to more significant improvements.
Given the lack of validation split in most benchmarks we experimented with, we do not include the result of tuning $\alpha$ as our main results.
However, for practitioners who have access to validation data, we recommend that they can grid search over $\alpha$ in case tuning leads to additional gain.

\subsection{Alternate Formulation}
\label{sec:app-alternate-formulation}
\begin{table}[]
\footnotesize
\centering
\begin{tabular}{lll}
\toprule
{} &               HumanEval &                  MBPP-S \\
\midrule
Coder-Reviewer       &  $56.7$ &           $\mathbf{64.4}$ \\
N. Coder-Reviewer &  $\mathbf{62.5}$ &  $61.6$ \\
\midrule
Alternate & $55.30$ & $63.0$\\
N. Alternate & $50.73$ & $62.4$\\
\bottomrule
\end{tabular}

\caption{Ranking results of the alternate formulation of Maximum Mutual Information (\Cref{sec:app-alternate-formulation}), which we observe underperforms Coder-Reviewer.
}
\label{tab:alternate_table}
\vspace{-10pt}
\end{table}
In \Cref{sec:coder-reviewer}, we show that Coder-Reviewer is related to pointwise mutual information regularization, when the regularization strength $\alpha$ is set to $0.5$.
There is another formulation of this objective that is also common in literature.
With derivation shown in \Cref{sec:app-derivation}, we can observe that 
\begin{align*}
    &\text{argmax}_{y}\;(1-\alpha)\log p(y|x) + \alpha \log p(x|y)\\
    =&\;\text{argmax}_{y}\;\log p(y|x) - \alpha \log p(y).
\end{align*}
Here we compare to this alternate formulation (\underline{Alternate}) on the 0-shot Python function completion datasets.
We use prompting to estimate $p(y)$ by removing the docstring beneath the function header and record the probability of the function body.
To apply this formulation with length normalization, we modify the regularization term as $\frac{1}{\lvert y \rvert} \log p(y)$.
From \Cref{tab:alternate_table} we observe that the alternate formulation generally underperforms Coder-Reviewer.

\begin{table*}[]
\setlength{\tabcolsep}{1.5pt}
\footnotesize
\centering
\begin{tabular}{l|lccccccccc}
\toprule
    Exec. Filter & Method &            Codex002 &            Codex001 &       Codex-Cushman &        InCoder6B &           InCoder1B &          CodeGen16B &           CodeGen6B &           CodeGen2B \\
\midrule
\multirow{6}{*}{No} & Random &              $53.7$ &              $34.7$ &              $38.8$ &           $19.9$ &              $11.6$ &              $28.1$ &              $20.4$ &              $20.8$ \\
    & Reviewer &              $60.7$ &              $53.0$ &              $47.5$ &  $\mathbf{35.5}$ &              $22.3$ &              $40.2$ &  $\underline{31.0}$ &              $25.7$ \\
\cmidrule{2-10}
    & Coder &              $55.4$ &              $47.8$ &              $43.2$ &           $14.4$ &               $8.8$ &              $31.0$ &              $21.9$ &              $23.2$ \\
    & Coder-Reviewer &              $63.0$ &              $54.2$ &              $50.3$ &           $33.7$ &              $21.6$ &              $40.3$ &              $25.9$ &              $25.9$ \\
\cmidrule{2-10}
    & N.Coder &              $57.1$ &              $44.8$ &              $39.4$ &           $20.8$ &               $9.6$ &              $28.9$ &              $21.8$ &              $23.6$ \\
    & N.Coder-Reviewer &              $64.9$ &              $53.4$ &              $49.6$ &           $32.8$ &              $21.5$ &              $41.0$ &              $27.7$ &              $24.5$ \\
\midrule
\multirow{7}{*}{Yes} & Random &              $57.6$ &              $46.8$ &              $41.0$ &           $21.9$ &              $13.4$ &              $36.7$ &              $25.8$ &              $28.9$ \\
    & Reviewer &  $\underline{65.2}$ &  $\underline{57.1}$ &  $\underline{50.9}$ &  $\mathbf{35.5}$ &     $\mathbf{30.3}$ &     $\mathbf{46.5}$ &     $\mathbf{32.6}$ &     $\mathbf{32.9}$ \\
\cmidrule{2-10}
    & Coder &              $57.7$ &              $51.1$ &              $45.8$ &           $15.5$ &               $8.9$ &              $38.6$ &              $25.6$ &              $30.1$ \\
    & Coder-Reviewer &              $58.3$ &              $51.9$ &              $46.0$ &           $16.4$ &              $10.7$ &              $41.7$ &              $25.4$ &              $30.1$ \\
\cmidrule{2-10}
    & N.Coder &              $59.4$ &              $47.8$ &              $41.8$ &           $22.6$ &              $11.0$ &              $37.4$ &              $26.3$ &              $28.8$ \\
    & N.Coder-Reviewer &     $\mathbf{68.0}$ &     $\mathbf{57.1}$ &     $\mathbf{52.0}$ &           $32.8$ &  $\underline{27.9}$ &  $\underline{45.9}$ &              $30.4$ &  $\underline{30.7}$ \\
\cmidrule{2-10}
    & MBR-\textsc{Exec} &              $60.9$ &              $46.6$ &              $46.4$ &           $21.0$ &               $8.1$ &              $37.0$ &              $25.3$ &              $30.0$ \\
\bottomrule
\end{tabular}

\caption{
Ranking results on the Plotting dataset. We observe that N.Coder-Reviewer and Reviewer alternate to be the best performing method. Executability filtering improves most methods and does not change the comparison between methods.
}
\label{tab:app_plotting_exec}
\end{table*}

\begin{table*}[]
\setlength{\tabcolsep}{1.5pt}
\footnotesize
\centering
\begin{tabular}{l|lccccccccc}
\toprule
    Exec. Filter & Method &            Codex002 &            Codex001 &       Codex-Cushman &           InCoder6B &           InCoder1B &          CodeGen16B &           CodeGen6B &           CodeGen2B \\
\midrule
\multirow{6}{*}{No} & Random &              $44.0$ &              $34.6$ &              $32.5$ &              $14.1$ &               $8.2$ &              $30.2$ &              $25.1$ &              $22.6$ \\
    & Reviewer &              $53.4$ &              $42.9$ &              $36.0$ &              $14.7$ &               $8.4$ &              $35.6$ &              $29.9$ &              $23.3$ \\
\cmidrule{2-10}
    & Coder &              $38.7$ &              $29.0$ &              $30.4$ &              $18.5$ &              $10.5$ &              $31.6$ &              $27.5$ &              $25.1$ \\
    & Coder-Reviewer &              $50.6$ &              $41.6$ &              $33.9$ &              $21.1$ &              $10.6$ &              $37.3$ &              $31.0$ &              $27.2$ \\
\cmidrule{2-10}
    & N.Coder &              $56.5$ &              $44.2$ &              $39.7$ &              $17.6$ &               $9.2$ &              $36.1$ &              $29.8$ &              $26.0$ \\
    & N.Coder-Reviewer &              $57.8$ &  $\underline{49.5}$ &              $39.7$ &              $20.1$ &              $10.5$ &              $37.8$ &              $31.3$ &              $26.1$ \\
\midrule
\multirow{7}{*}{Yes} & Random &              $49.2$ &              $38.1$ &              $35.1$ &              $16.0$ &               $9.5$ &              $32.2$ &              $26.4$ &              $24.2$ \\
    & Reviewer &  $\underline{61.2}$ &              $47.0$ &              $41.3$ &              $17.6$ &               $9.5$ &              $37.3$ &              $31.3$ &              $25.2$ \\
\cmidrule{2-10}
    & Coder &              $45.1$ &              $32.7$ &              $33.0$ &              $20.1$ &              $10.8$ &              $33.5$ &              $28.3$ &              $26.3$ \\
    & Coder-Reviewer &              $56.7$ &              $46.2$ &              $39.1$ &     $\mathbf{23.5}$ &     $\mathbf{12.0}$ &     $\mathbf{40.0}$ &     $\mathbf{32.5}$ &     $\mathbf{28.5}$ \\
\cmidrule{2-10}
    & N.Coder &              $60.2$ &              $47.2$ &  $\underline{42.1}$ &              $19.2$ &              $10.2$ &              $38.5$ &              $30.9$ &              $26.6$ \\
    & N.Coder-Reviewer &     $\mathbf{62.5}$ &     $\mathbf{52.1}$ &     $\mathbf{43.8}$ &  $\underline{22.0}$ &  $\underline{11.5}$ &  $\underline{39.6}$ &  $\underline{32.2}$ &  $\underline{27.4}$ \\
\cmidrule{2-10}
    & MBR-\textsc{Exec} &              $50.5$ &              $36.5$ &              $35.6$ &              $20.7$ &              $10.9$ &              $35.8$ &              $30.6$ &              $27.2$ \\
\bottomrule
\end{tabular}

\caption{
Ranking results on the HumanEval dataset. We observe that N.Coder-Reviewer works the best on Codex model families and Coder-Reviewer works the best on CodeGen and Incoder models. Executability filtering improves most methods and does not change the comparison between methods.
}
\label{tab:app_humaneval_exec}
\end{table*}

\begin{table*}[]
\setlength{\tabcolsep}{1.5pt}
\footnotesize
\centering
\begin{tabular}{l|lccccccccc}
\toprule
    Exec. Filter & Method &            Codex002 &            Codex001 &       Codex-Cushman &           InCoder6B &           InCoder1B &          CodeGen16B &           CodeGen6B &           CodeGen2B \\
\midrule
\multirow{6}{*}{No} & Random &              $55.6$ &              $50.0$ &              $44.0$ &              $22.6$ &              $16.7$ &              $43.5$ &              $40.1$ &              $34.1$ \\
    & Reviewer &              $56.3$ &              $51.0$ &              $45.4$ &              $25.6$ &              $17.7$ &              $42.0$ &              $41.3$ &              $38.3$ \\
\cmidrule{2-10}
    & Coder &              $55.9$ &              $52.3$ &              $44.7$ &              $26.0$ &              $20.8$ &              $49.2$ &              $45.1$ &              $36.9$ \\
    & Coder-Reviewer &              $61.5$ &              $56.9$ &              $51.0$ &  $\underline{31.7}$ &              $23.4$ &              $46.1$ &              $44.9$ &              $41.5$ \\
\cmidrule{2-10}
    & N.Coder &              $58.3$ &              $53.6$ &              $48.7$ &              $24.8$ &              $21.0$ &              $47.5$ &              $44.0$ &              $38.2$ \\
    & N.Coder-Reviewer &              $59.6$ &              $55.6$ &              $51.6$ &              $28.8$ &              $20.0$ &              $44.4$ &              $43.2$ &              $40.1$ \\
\midrule
\multirow{7}{*}{Yes} & Random &              $58.8$ &              $53.3$ &              $46.7$ &              $25.4$ &              $19.3$ &              $45.7$ &              $42.7$ &              $36.5$ \\
    & Reviewer &              $59.5$ &              $54.7$ &              $48.2$ &              $28.8$ &              $21.0$ &              $43.6$ &              $43.5$ &              $40.3$ \\
\cmidrule{2-10}
    & Coder &              $59.8$ &              $55.9$ &              $47.5$ &              $28.1$ &              $23.4$ &  $\underline{50.3}$ &  $\underline{47.3}$ &              $39.3$ \\
    & Coder-Reviewer &     $\mathbf{64.4}$ &     $\mathbf{60.3}$ &     $\mathbf{53.9}$ &     $\mathbf{33.7}$ &  $\underline{25.8}$ &              $47.4$ &              $47.1$ &  $\underline{43.3}$ \\
\cmidrule{2-10}
    & N.Coder &              $60.5$ &              $55.5$ &              $49.8$ &              $27.1$ &              $22.6$ &              $48.6$ &              $45.3$ &              $40.3$ \\
    & N.Coder-Reviewer &              $61.6$ &              $57.7$ &  $\underline{53.1}$ &              $31.1$ &              $22.6$ &              $45.8$ &              $44.7$ &              $41.9$ \\
\cmidrule{2-10}
    & MBR-\textsc{Exec} &  $\underline{63.9}$ &  $\underline{59.4}$ &              $50.7$ &              $30.9$ &     $\mathbf{26.2}$ &     $\mathbf{53.5}$ &     $\mathbf{48.3}$ &     $\mathbf{43.7}$ \\
\bottomrule
\end{tabular}

\caption{
Ranking results on the MBPP-S dataset. We observe that Coder-Reviewer works the best on Codex model families and MBR-\textsc{Exec} is usually the best on CodeGen and Incoder models. Executability filtering improves most methods and usually does not change the comparison between methods.
}
\label{tab:app_mbpps_exec}
\end{table*}

\begin{table*}[]
\setlength{\tabcolsep}{1.5pt}
\footnotesize
\centering
\begin{tabular}{l|lccccccccc}
\toprule
    Exec. Filter & Method &            Codex002 &            Codex001 &       Codex-Cushman &           InCoder6B &           InCoder1B &          CodeGen16B &           CodeGen6B &           CodeGen2B \\
\midrule
\multirow{6}{*}{No} & Random &              $53.6$ &              $46.9$ &              $35.1$ &              $14.8$ &               $9.1$ &              $33.5$ &              $28.5$ &              $24.1$ \\
    & Reviewer &              $63.3$ &              $53.1$ &              $41.5$ &              $20.8$ &              $13.2$ &              $40.5$ &              $32.7$ &              $28.4$ \\
\cmidrule{2-10}
    & Coder &              $55.4$ &              $49.8$ &              $38.8$ &              $19.7$ &              $12.3$ &              $36.1$ &              $32.7$ &              $28.9$ \\
    & Coder-Reviewer &              $62.6$ &              $55.2$ &              $45.0$ &              $23.3$ &              $14.8$ &              $42.2$ &              $34.6$ &              $30.5$ \\
\cmidrule{2-10}
    & N.Coder &              $55.5$ &              $50.6$ &              $35.1$ &              $15.9$ &               $9.3$ &              $36.9$ &              $32.4$ &              $27.9$ \\
    & N.Coder-Reviewer &              $62.5$ &              $54.5$ &              $42.2$ &              $20.9$ &              $13.2$ &              $41.4$ &              $33.4$ &              $29.3$ \\
\midrule
\multirow{7}{*}{Yes} & Random &              $58.1$ &              $51.6$ &              $40.8$ &              $19.6$ &              $13.1$ &              $39.3$ &              $34.2$ &              $28.2$ \\
    & Reviewer &     $\mathbf{66.9}$ &              $57.4$ &              $46.7$ &              $24.4$ &              $15.7$ &              $44.1$ &              $39.4$ &              $31.7$ \\
\cmidrule{2-10}
    & Coder &              $60.2$ &              $55.1$ &              $44.7$ &              $23.4$ &              $16.0$ &              $41.5$ &              $37.9$ &              $31.9$ \\
    & Coder-Reviewer &  $\underline{66.4}$ &     $\mathbf{59.4}$ &     $\mathbf{48.9}$ &  $\underline{26.1}$ &  $\underline{17.1}$ &  $\underline{46.2}$ &  $\underline{40.3}$ &  $\underline{33.2}$ \\
\cmidrule{2-10}
    & N.Coder &              $60.7$ &              $54.8$ &              $40.8$ &              $20.2$ &              $13.3$ &              $41.9$ &              $37.5$ &              $31.0$ \\
    & N.Coder-Reviewer &              $66.2$ &              $58.4$ &              $46.3$ &              $24.1$ &              $15.8$ &              $45.4$ &              $39.8$ &              $32.3$ \\
\cmidrule{2-10}
    & MBR-\textsc{Exec} &              $63.0$ &  $\underline{58.6}$ &  $\underline{48.3}$ &     $\mathbf{26.7}$ &     $\mathbf{18.3}$ &     $\mathbf{47.3}$ &     $\mathbf{41.1}$ &     $\mathbf{35.5}$ \\
\bottomrule
\end{tabular}

\caption{
Ranking results on the MBPP dataset. We observe that Coder-Reviewer usually works the best on Codex model families; MBR-\textsc{Exec} is usually the best on CodeGen and Incoder models and Coder-Reviewer is usually the second best performing method. Executability filtering improves most methods and usually does not change the comparison between methods.
}
\label{tab:app_mbpp_exec}
\end{table*}

\begin{table*}[]
\setlength{\tabcolsep}{1.5pt}
\footnotesize
\centering
\begin{tabular}{l|lccccccccc}
\toprule
    Exec. Filter & Method &            Codex002 &            Codex001 &       Codex-Cushman &           InCoder6B &           InCoder1B &          CodeGen16B &           CodeGen6B &           CodeGen2B \\
\midrule
\multirow{6}{*}{No} & Random &              $60.0$ &              $55.2$ &              $55.2$ &              $49.7$ &              $41.0$ &              $35.6$ &              $33.2$ &              $25.4$ \\
    & Reviewer &              $63.3$ &              $58.4$ &              $59.9$ &              $55.4$ &              $43.3$ &              $28.3$ &              $29.4$ &              $19.5$ \\
\cmidrule{2-10}
    & Coder &              $57.4$ &              $48.0$ &              $50.8$ &              $48.7$ &              $42.6$ &              $31.3$ &              $25.2$ &              $25.6$ \\
    & Coder-Reviewer &              $61.9$ &              $57.7$ &              $58.4$ &              $55.0$ &  $\underline{50.8}$ &  $\underline{37.3}$ &     $\mathbf{40.3}$ &     $\mathbf{37.0}$ \\
\cmidrule{2-10}
    & N.Coder &              $61.0$ &              $49.9$ &              $48.2$ &              $40.8$ &              $37.5$ &              $36.8$ &              $32.9$ &              $25.4$ \\
    & N.Coder-Reviewer &     $\mathbf{63.7}$ &  $\underline{59.9}$ &              $60.0$ &     $\mathbf{56.0}$ &              $42.8$ &              $29.1$ &              $31.0$ &              $20.8$ \\
\midrule
\multirow{7}{*}{Yes} & Random &              $60.0$ &              $55.2$ &              $55.3$ &              $49.8$ &              $41.1$ &              $35.7$ &              $33.2$ &              $25.7$ \\
    & Reviewer &              $63.3$ &              $58.5$ &  $\underline{60.2}$ &              $55.3$ &              $43.3$ &              $28.1$ &              $29.2$ &              $19.9$ \\
\cmidrule{2-10}
    & Coder &              $57.4$ &              $48.1$ &              $50.9$ &              $48.8$ &              $42.6$ &              $31.7$ &              $25.4$ &              $25.9$ \\
    & Coder-Reviewer &              $61.9$ &              $57.8$ &              $58.5$ &              $55.0$ &     $\mathbf{50.9}$ &              $37.0$ &  $\underline{39.8}$ &  $\underline{36.7}$ \\
\cmidrule{2-10}
    & N.Coder &              $61.3$ &              $49.9$ &              $48.8$ &              $41.5$ &              $38.0$ &     $\mathbf{37.8}$ &              $33.1$ &              $25.6$ \\
    & N.Coder-Reviewer &  $\underline{63.7}$ &     $\mathbf{59.9}$ &     $\mathbf{60.4}$ &  $\underline{55.9}$ &              $43.1$ &              $29.5$ &              $30.8$ &              $21.1$ \\
\cmidrule{2-10}
    & MBR-\textsc{Exec} &              $57.4$ &              $48.3$ &              $51.0$ &              $48.8$ &              $42.6$ &              $32.4$ &              $25.8$ &              $26.3$ \\
\bottomrule
\end{tabular}

\caption{
Ranking results on the NL2Bash dataset. We observe that Coder-Reviewer works the best on Codex model families and MBR-\textsc{Exec} is usually the best on CodeGen and Incoder models. Executability filtering is implemented with simulated execution via parsing the generated bash code and does not lead to a consistent improvement.
}
\label{tab:app_nl2bash_exec}
\end{table*}

\begin{table*}[]
\setlength{\tabcolsep}{1.5pt}
\footnotesize
\centering
\begin{tabular}{l|lccccccccc}
\toprule
    Exec. Filter & Method &            Codex002 &            Codex001 &       Codex-Cushman &           InCoder6B &           InCoder1B &          CodeGen16B &           CodeGen6B &           CodeGen2B \\
\midrule
\multirow{6}{*}{No} & Random &              $62.8$ &              $46.0$ &              $40.2$ &              $19.0$ &              $11.7$ &              $13.6$ &              $21.1$ &              $11.9$ \\
    & Reviewer &              $62.9$ &              $50.0$ &              $44.2$ &              $27.3$ &              $16.8$ &              $16.0$ &              $22.4$ &              $13.1$ \\
\cmidrule{2-10}
    & Coder &              $71.5$ &              $56.8$ &              $51.8$ &              $32.4$ &              $21.4$ &              $27.6$ &              $30.4$ &              $21.5$ \\
    & Coder-Reviewer &              $71.6$ &              $56.5$ &              $52.8$ &              $32.8$ &              $22.1$ &              $23.8$ &              $28.6$ &              $17.4$ \\
\cmidrule{2-10}
    & N.Coder &              $67.4$ &              $56.5$ &              $44.0$ &              $29.5$ &              $17.2$ &              $22.4$ &              $26.9$ &              $17.0$ \\
    & N.Coder-Reviewer &              $68.0$ &              $54.7$ &              $48.3$ &              $30.7$ &              $20.6$ &              $18.9$ &              $24.8$ &              $14.9$ \\
\midrule
\multirow{7}{*}{Yes} & Random &              $65.2$ &              $57.1$ &              $50.8$ &              $29.4$ &              $18.8$ &              $25.6$ &              $32.7$ &              $22.4$ \\
    & Reviewer &              $67.5$ &              $60.0$ &              $54.7$ &              $38.4$ &              $26.8$ &              $28.8$ &              $36.8$ &              $23.1$ \\
\cmidrule{2-10}
    & Coder &              $74.1$ &  $\underline{64.3}$ &              $58.8$ &              $38.9$ &              $26.9$ &     $\mathbf{33.7}$ &  $\underline{37.3}$ &     $\mathbf{27.5}$ \\
    & Coder-Reviewer &  $\underline{74.5}$ &     $\mathbf{65.3}$ &     $\mathbf{60.4}$ &     $\mathbf{41.5}$ &     $\mathbf{29.8}$ &  $\underline{31.7}$ &     $\mathbf{38.8}$ &              $24.8$ \\
\cmidrule{2-10}
    & N.Coder &              $69.9$ &              $63.2$ &              $54.1$ &              $38.2$ &              $26.8$ &              $31.1$ &              $36.8$ &              $25.7$ \\
    & N.Coder-Reviewer &              $71.0$ &              $62.6$ &              $57.6$ &  $\underline{40.3}$ &  $\underline{29.0}$ &              $29.9$ &  $\underline{37.3}$ &              $23.6$ \\
\cmidrule{2-10}
    & MBR-\textsc{Exec} &     $\mathbf{75.2}$ &              $63.2$ &  $\underline{59.0}$ &              $38.2$ &              $27.3$ &              $30.6$ &              $37.0$ &  $\underline{26.1}$ \\
\bottomrule
\end{tabular}

\caption{
Ranking results on the Spider dataset. We observe that Coder-Reviewer usually works the best. Executability filtering improves most methods and usually does not change the comparison between methods.
}
\label{tab:app_spider_exec}
\end{table*}

\end{document}